\documentclass[conference]{IEEEtran}
\usepackage{times}

\usepackage[numbers]{natbib}
\usepackage{multicol}
\usepackage[bookmarks=true]{hyperref}
\usepackage[english]{babel}
\usepackage{amsmath,amssymb,amsfonts}
\usepackage{algorithmic}
\usepackage{graphicx}
\usepackage{textcomp}
\usepackage{xcolor}
\usepackage{tabularx}
\usepackage{xspace}
\usepackage{url}
\usepackage{booktabs}
\usepackage{threeparttable}
\usepackage{array}
\usepackage{caption}
\usepackage{pifont}
\usepackage{bm}
\let\labelindent\relax
\usepackage{enumitem}
\usepackage{gensymb}
\usepackage{lipsum}
\usepackage{float}
\usepackage{tcolorbox}

\definecolor{royalblue}{RGB}{65, 105, 225}
\definecolor{softgreen}{RGB}{85, 170, 85} 
\definecolor{softred}{RGB}{200, 50, 50}   

\hypersetup{
    colorlinks=true,
    linkcolor=orange,    
    citecolor=softgreen,    
    urlcolor=royalblue      
}

\def\BibTeX{{\rm B\kern-.05em{\sc i\kern-.025em b}\kern-.08em
    T\kern-.1667em\lower.7ex\hbox{E}\kern-.125emX}}

\newcommand{\system}{Minimalist Compliance Control\xspace}

\author{
\authorblockN{ 
\textbf{Haochen Shi\authorrefmark{1}} \quad
\textbf{Songbo Hu\authorrefmark{1}} \quad
\textbf{Yifan Hou} \quad
\textbf{Weizhuo Wang} \quad
\textbf{C. Karen Liu\authorrefmark{2}} \quad
\textbf{Shuran Song\authorrefmark{2}}
}
\vspace{1mm}
\authorblockA{
\authorrefmark{1}Equal contribution \quad
\authorrefmark{2}Equal advising
}
\vspace{1mm}
\authorblockA{Stanford University}
\vspace{1mm}
\authorblockA{
\textbf{\textcolor{magenta}{\url{https://minimalist-compliance-control.github.io}}}
\vspace{-2mm}
}
}

\begin{document}

\title{Minimalist Compliance Control}

\twocolumn[{%
\renewcommand\twocolumn[1][]{#1}%
\maketitle
\includegraphics[width=\textwidth]{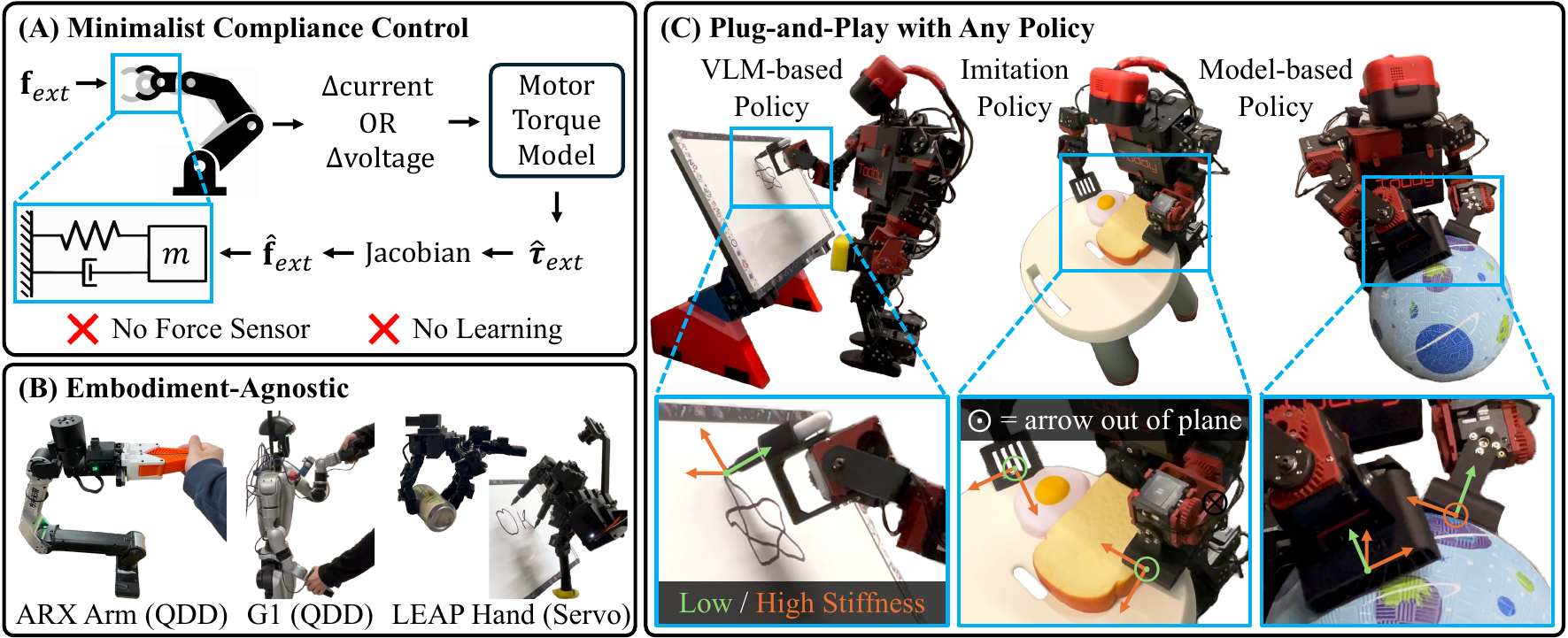}
\vspace{-4mm}
\captionof{figure}{
\textbf{\system} (A) requires no force sensors or learning, estimating external wrenches $\hat{\mathbf{f}}_{\text{ext}}$ directly from motor current or voltage signals using a motor torque model and Jacobians. These estimates drive a spring--mass--damper model to update task-space position references. (B) This minimalist approach generalizes across embodiment, from robot arm and dexterous hand to humanoid robot, and (C) remains plug-and-play with any policy such as VLM-based policy, imitation policy, and model-based policy, and across diverse tasks such as wiping, drawing, scooping, and in-hand manipulation.}
\vspace{3mm}
\label{fig:teaser}
}]

\begin{abstract}
Compliance control is essential for safe physical interaction, yet its adoption is limited by hardware requirements such as force/torque sensors. While recent reinforcement learning approaches aim to bypass these constraints, they often suffer from sim-to-real gaps, lack safety guarantees, and add system complexity. We propose Minimalist Compliance Control, which enables compliant behavior using only motor current or voltage signals readily available in modern servos and quasi-direct-drive motors—without force sensors, current control, or learning. External wrenches are estimated from actuator signals and Jacobians and incorporated into a task-space admittance controller, preserving sufficient force measurement accuracy for stable and responsive compliance control. Our method is embodiment-agnostic and plug-and-play with diverse high-level planners. We validate our approach on a robot arm, a dexterous hand, and two humanoid robots across multiple contact-rich tasks, using vision--language models, imitation learning, and model-based planning. The results demonstrate robust, safe, and compliant interaction across embodiments and planning paradigms.
\end{abstract}
\section{Introduction}

Widespread adoption of compliance control is currently hindered by significant hardware barriers. Standard admittance and impedance control typically rely on expensive force/torque sensors or actuators with precise force feedback capabilities that are not available on many widely used robot platforms~\cite{craig1979systematic,mason1981compliance,khatib1987unified,martin-martin2019variable,zhang2021learning,hou2025adaptivea,xu2025compliant,wang2022safe,zhang2023efficient,choi2026inthewild}.
To address this, recent work has proposed leveraging Reinforcement Learning (RL) to learn compliance control policies~\citep{wei2025hmc,zhi2025learning,xu2025faceta,lu2025gentlehumanoid,chen2025chip,he2025cotap,zhou2025hacloco,margolis2025softmimic}. Typically, these approaches learn a position controller to mimic compliance behavior through reward shaping during training. However, there are two major limitations of RL-based compliance:
First, these methods suffer from the sim-to-real gap when deployed in the real world, such as discrepancies in position tracking and stiffness accuracy. Consequently, the learned controllers often lack safety guarantees and can produce unexpected, large force spikes during physical interaction~\citep{zhi2025learning,xu2025faceta,lu2025gentlehumanoid}. 
Moreover, RL-based pipelines increase the framework's complexity and can be difficult to tune~\citep{margolis2025softmimic,wei2025hmc}. 

To overcome hardware barriers while avoiding unnecessary complexity, a key observation is that motor current or PWM (pulse-width modulation, which is essentially voltage) inherently contains information for estimating external wrenches. In fact, early work in the 1990s leveraged motor current for torque estimation, but required carefully designed observer-based filtering to mitigate substantial measurement noise and recover physically meaningful external forces~\citep{murakami1993torque,eom1998disturbance}. In contrast, modern actuator designs feature significantly improved current sensing and high-bandwidth current control loops, and have demonstrated that torque estimated from motor current closely matches ground-truth measurements~\citep{seok2013design,hutter2016anymal,yu2020quasidirect}. 

Another key observation is that inaccurate stiffness realization and imprecise external wrench estimation affect compliance in different ways. Compliance control does not require highly accurate force magnitudes; instead, it primarily depends on correct force sign/direction and reasonable accuracy in the relevant frequency band to ensure stable and timely responses to disturbances. When these properties are preserved, approximate wrench estimates are sufficient to induce compliant behavior, and improved magnitude accuracy mainly enhances performance rather than safety~\citep{zhi2025learning}. 
In contrast, RL-based approaches do not explicitly guarantee either force sign consistency or frequency-domain behavior.

Therefore, we propose Minimalist Compliance Control for robots equipped with modern servos or quasi-direct-drive (QDD) motors (Fig.~\ref{fig:teaser}). Our method (1) calibrates the motor characteristics (e.g., torque constant), (2) estimates external wrenches from motor current or PWM, and (3) applies task-space admittance control using the estimated wrench. This approach provides three advantages that together democratize compliant behavior for a broad class of robotic systems:

\begin{itemize}
    \item \textbf{No Force Sensors:} By estimating external forces and torques solely via motor current or PWM signals ubiquitous in modern actuators, we eliminate the dependency on direct force/torque/tactile sensing hardware, drastically reducing system cost and maintenance requirements. Notably, our approach does not even require current sensors or current-controlled motors; raw PWM signals alone are sufficient. Despite its simplicity, this model-based estimation preserves sufficient force measurement accuracy for stable and responsive compliance control. In contrast, RL-based methods provide no explicit guarantees on these, increasing the risk of unsafe behavior during real-world contact. Meanwhile, our approach retains additional advantages, including reactive behavior and the ability to respond to disturbances beyond end-effector wrenches. 
    \item \textbf{Embodiment-Agnostic:} Grounded in task-space admittance control, our framework naturally generalizes across diverse morphologies, from fixed-base robot arms to dexterous hands and floating-base humanoids.
    \item \textbf{Plug-and-Play with Any Policy:} Our model-based approach remains compatible with a wide variety of high-level planning strategies and requires minimal tuning, ranging from vision--language models (VLMs), imitation learning, to model-based motion planning.
\end{itemize}

In summary, we present a Minimalist Compliance Control framework that enables compliant interaction without force sensors or learning, extending this capability to a wide range of robotic systems equipped with modern servo and QDD motors. We validate our method through comprehensive experiments on four distinct platforms—a robot arm, a dexterous hand, and two humanoid robots—across contact-rich tasks such as wiping, drawing, scooping, and in-hand manipulation.




\section{Related Works}

\subsection{Robots and Force Torque Sensors}
Many widely used robot platforms lack end-effector force/torque sensors or joint torque sensors. This is a limitation that spans diverse embodiments, including robot arms (e.g., ARX, ViperX, and ALOHA~\citep{aldaco2024aloha}), dexterous hands (e.g., LEAP~\citep{shaw2023leap}, ORCA~\citep{christoph2025orca}, and DexHand~\citep{yuan2025development}), and humanoid robots (e.g., Unitree G1, Berkeley Humanoid Lite~\cite{chi2025demonstrating}, and ToddlerBot~\cite{shi2025toddlerbot,hu2025robot,yang2026locomotion}). While installing external 6-axis force/torque sensors offers a direct solution, industry-standard options like the ATI Mini-45 are often bulky, limited in robustness, and prohibitively expensive~\citep{stassi2014flexible,cao2021sixaxis,choi2025coinft}. While alternative tactile technologies like optical~\citep{yuan2017gelsight}, resistive~\citep{huang20243dvitac}, and magnetic~\citep{bhirangi2025anyskin} sensors exist, they are often limited by durability, signal drift, or susceptibility to interference. Moreover, they generally require complex calibration and primarily sense normal forces and rough estimates of shear forces. Another key observation is that these robotic systems are all equipped with modern servo and QDD motors, which motivates our approach to estimate external wrenches from motor current or PWM signals that are readily available on these platforms.

\subsection{Model-based Compliance Control}
Classical model-based strategies, such as task-space admittance and impedance control~\citep{craig1979systematic, mason1981compliance, khatib1987unified}, provide principled frameworks for regulating contact forces via virtual spring–mass–damper dynamics. However, these approaches typically require either 6-axis force/torque sensors or accurate joint torque sensing and control.
To relax these hardware requirements, prior work has explored sensorless force estimation using joint-level disturbance observers that separate externally induced torques from internally modeled dynamics, enabling force or motion control without dedicated force sensors~\citep{murakami1993torque,eom1998disturbance}. However, due to limited sensing accuracy and edge computing in the 1990s, current measurements were often extremely noisy and required extensive filtering, ultimately limiting their practical usefulness. 
More recently, improved current sensing, high-bandwidth current control, and proper calibration have enabled accurate torque estimation directly from motor current, primarily in systems with low transmission distortion such as direct-drive and QDD motors~\citep{seok2013design,hutter2016anymal,yu2020quasidirect}. Our work extends beyond low-transmission, high–torque-transparency actuators and shows that servo motors with high-reduction transmissions (gear ratios exceeding $200{:}1$), even without current/torque control, can be reliably used for torque estimation and subsequent admittance control.


\subsection{RL-based Compliance Control}
As another approach to address hardware limitations, recent work has explored reinforcement learning (RL) for compliant control, aiming to implicitly infer external wrenches from proprioceptive signals through a neural network~\citep{margolis2025softmimic,xu2025faceta,zhi2025learning}. These methods generally fall into two categories: task-specific policies, which train on augmented datasets around specific motion clips to absorb disturbances~\citep{margolis2025softmimic}, and task-agnostic policies, which aim to generalize across tasks by mimicking spring--mass--damper dynamics~\citep{zhi2025learning,xu2025faceta,lu2025gentlehumanoid}. However, the black-box nature of these policies often leads to sim-to-real discrepancies, resulting in a lack of safety guarantees and susceptibility to dangerous force spikes. 
In this work, we achieve reliable compliance control using explicit wrench estimation and classical model-based control, without sim-to-real transfer or black-box policies.


\section{Method}

We present our low-level compliance controller in Section~\ref{sec:model},~\ref{sec:motor},~\ref{sec:wrench},~\ref{sec:admittance} and high-level planning strategies in Section~\ref{sec:vlm},~\ref{sec:imitation},~\ref{sec:model_based}. Each planning strategy can be used independently with our compliance controller. Fig.~\ref{fig:planning} summarizes the inputs and outputs of each policy.

\subsection{Spring--Mass--Damper Model}
\label{sec:model}

In our compliance controller, each end-effector is modeled as a spring--mass--damper system that responds to commanded motion and external wrenches. The dynamics are governed by:
\begin{equation}
    m \ddot{\mathbf{x}} = \mathbf{K}_p (\mathbf{x}_{\text{des}} - \mathbf{x}) + \mathbf{K}_d (\dot{\mathbf{x}}_{\text{des}} - \dot{\mathbf{x}}) + \mathbf{f}_{\text{cmd}} + \mathbf{f}_{\text{ext}}, \label{eq:smd}
\end{equation}
where $\mathbf{x} \in \mathbb{R}^6$ denotes the end-effector pose with orientation represented as a rotation vector, $m$ is the effective mass (set to 1 for simplicity), $\mathbf{K}_p$ and $\mathbf{K}_d$ are the stiffness and damping matrices, $\mathbf{x}_{\text{des}}$ and $\dot{\mathbf{x}}_{\text{des}}$ are the desired pose and velocity, and $\mathbf{f}_{\text{cmd}}$ and $\mathbf{f}_{\text{ext}}$ are the commanded and external wrenches. 

\subsection{Motor Torque Estimation}
\label{sec:motor}

\begin{figure*}[t]
  \centering
  \includegraphics[width=\textwidth]{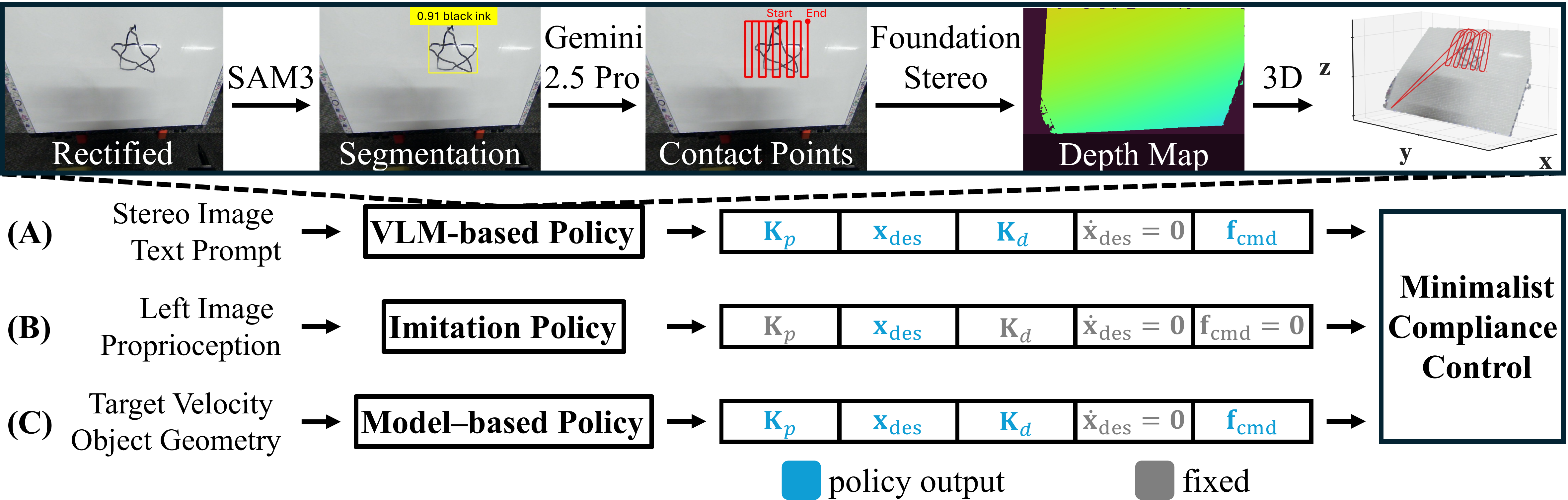}
  \caption{\textbf{Policy Inputs and Outputs.} We show that Minimalist Compliance Control is plug-and-play with a range of high-level policies that benefit from compliant interaction. We illustrate the inputs and outputs of (A) a VLM-based policy, (B) an imitation policy, and (C) a model-based policy. The math symbols refer to those in Eq.~\eqref{eq:smd}. (A) and (C) predict only the directions of the stiffness $\mathbf{K}_p$ and force command $\mathbf{f}_{\text{cmd}}$, while their magnitudes remain fixed. For all three, the damping matrix is set for critical damping as $\mathbf{K}_d = 2\,\mathbf{K}_p^{1/2}$, assuming an identity inertia matrix, where $(\cdot)^{1/2}$ denotes the matrix square root. We also visualize the VLM pipeline to predict 3D contact points and normals, which are interpolated to generate $\mathbf{x}_{\text{des}}$.}
  \vspace{-3mm}
  \label{fig:planning}
\end{figure*}

Motor torque sensing is not directly available in most servo and QDD motors. Therefore, we estimate motor torques from motor current or pulse-width modulation (PWM) signals that can be used to infer motor current. We sysID the motors following standard motor models described in~\citep{gamazo-real2010position,lee2023how}.

For motors with current sensors, we directly read the motor current $I_{\text{w}}$ from the sensor.
For motors without current sensors, we estimate the motor current using the following relations:
\begin{align}
    V_{\text{PWM}} &= \text{PWM} \times V_{\text{bus}}, \label{eq:pwm_voltage} \\
    V_{\text{emf}} &= \frac{\dot{q}}{K_{\text{v}}}, \label{eq:back_emf} \\
    I_{\text{w}} &= \frac{V_{\text{PWM}} - V_{\text{emf}}}{R_{\text{w}}}, \label{eq:motor_current}
\end{align}
where $I_{\text{w}}$ is the motor winding current, $V_{\text{PWM}}$ is the PWM voltage, $V_{\text{emf}}$ is the back EMF, $R_{\text{w}}$ is the winding resistance, $\text{PWM} \in [0, 1]$ is the duty cycle, $V_{\text{bus}}$ is a constant bus voltage, $\dot{q}$ is the motor velocity, and $K_{\text{v}}$ is the velocity constant. 

With Eq.~\eqref{eq:back_emf}, we calibrate $K_{\text{v}}$ by sweeping the motor through different velocities under no-load conditions where $V_{\text{PWM}} \approx V_{\text{emf}}$. We estimate $R_{\text{w}}$ using an external current sensor to measure $I_{\text{w}}$ under varying PWM duty cycles, then solve for $R_{\text{w}}$ with Eq.~\eqref{eq:motor_current}. 
The motor torque constant $K_{\text{t}}$ is calibrated from the current-torque relationship:
\begin{equation}
    K_{\text{t}} = \frac{\tau_{\text{load}}}{\eta I_{\text{w}}}, \label{eq:torque_est}
\end{equation}
where $\tau_{\text{load}}$ is the measured load torque, $\eta \in (0,1]$ is the motor efficiency accounting for power conversion losses, and $I_{\text{w}}$ is the motor winding current. We use manufacturer-provided torque constants $K_{\text{t}}$ when available; otherwise, we calibrate $K_{\text{t}}$ under different load and velocity conditions. This means that for motors equipped with current sensors and manufacturer-provided $K_{\text{t}}$, no additional system identification is required.

To account for asymmetric power transmission in forward and backward drive, we apply direction-dependent gains:
\begin{align}
    d &= \begin{cases}
        \text{sign}(\tau_{\text{w}} \dot{q}), & |\dot{q}| > \epsilon_{\text{vel}}, \\
        d_{\text{prev}}, & |\dot{q}| \le \epsilon_{\text{vel}},
    \end{cases} \label{eq:drive_state} \\
    \tau_{\text{load}} &= \begin{cases}
        \eta K_{\text{t}} I_{\text{w}}, & d > 0 \text{ (forward drive)}, \\
        \eta^{-1} K_{\text{t}} I_{\text{w}}, & d \leq 0 \text{ (backward drive)},
    \end{cases} \label{eq:gain_select}
\end{align}
where $d$ is the forward/backward drive state determined from the sign of power flow $\tau_{\text{w}} \dot{q}$, $\epsilon_{\text{vel}}$ is a velocity threshold to debounce, and $\tau_{\text{load}}$ is the output torque at the joint. $d_{\text{prev}}$ is initialized to be 1. In forward drive ($d > 0$), the motor accelerates the load, and efficiency $\eta$ accounts for power losses. In backward drive ($d \leq 0$), external forces back-drive the motor, and the inverse efficiency $\eta^{-1}$ accounts for brake amplification through the transmission. The system identification is performed using measurements from the open-source motor testbed described in~\citep{shi2025toddlerbot}.

To isolate external torques, we compensate for the gravity by subtracting $\tau_{\text{grav}}$ from the measured torque:
\begin{align}
    \tau_{\text{ext}} &= -(r \cdot \tau_{\text{load}} - \tau_{\text{grav}}), \label{eq:ext_torque}
\end{align}
where $r$ is the gear ratio and $\tau_{\text{grav}}$ is the gravity-induced torque.  
We assume a quasi-static interaction regime, in which inertial and velocity-dependent terms—such as joint acceleration and Coriolis effects—are negligible compared to gravity and externally applied torques. 
Under this assumption, $\tau_{\text{ext}}$ provides a reliable estimate of the externally induced torque. 
Empirically, this assumption holds for most tasks that require compliance, where interactions are dominated by low-frequency, quasi-static contacts rather than highly dynamic motions.

\subsection{External Wrench Estimation}
\label{sec:wrench}


The external wrench is estimated from motor torques using a flexible formulation that supports either full-wrench recovery or selective, axis-aligned estimation.
When full-wrench estimation is enabled, we recover the contact force and optionally the torque by solving a regularized least-squares problem:
\begin{equation}
\hat{\mathbf{f}}_{\text{ext}}^{\,p}
=
\arg\min_{\mathbf{f}}
\;\big\lVert
J_p^\top \mathbf{f} - \bm{\tau}_{\text{ext}}
\big\rVert^2
+
\lambda \lVert \mathbf{f} \rVert^2,
\label{eq:full_wrench}
\end{equation}
where $\mathbf{f}_{\text{ext}}^{\,p}\in\mathbb{R}^3$ is the translational component of $\mathbf{f}_{\text{ext}}$, $J_p \in \mathbb{R}^{3\times n}$ is the translational Jacobian of the contact site, and $\lambda$ is a regularization coefficient.
Torque components can be recovered analogously using the rotational Jacobian $J_r$.

For improved numerical conditioning and noise robustness, we optionally estimate only the physically relevant components—namely, the force along the contact normal.
For a selected unit axis $\hat{\bm{u}}\in\mathbb{R}^3$, we estimate the corresponding axis-aligned force as
\begin{equation}
\hat{\mathbf{f}}_{\text{ext}}^{\,p}
=
\frac{(\hat{\bm{u}}^{T} J_p)\,\bm{\tau}_{\text{ext}}}
{(\hat{\bm{u}}^{T} J_p)(\hat{\bm{u}}^{T} J_p)^{T} + \lambda}
\;\hat{\bm{u}} .
\label{eq:axis_wrench}
\end{equation}
This formulation reduces to a set of independent 1D regularized least-squares problems, avoiding ill-conditioned Jacobian inversions while preserving compliant behavior in the directions of interest.
Torque components along selected axes are estimated in the same manner using $J_r$.

\subsection{Admittance Control with Inverse Kinematics}
\label{sec:admittance}

Given the estimated external wrench $\hat{\mathbf{f}}_{\text{ext}}$, admittance control is realized by integrating the spring--mass--damper dynamics to update the task-space motion reference. 
At each control step $k$, the reference acceleration $\ddot{\mathbf{x}}_{\text{ref}}^{k}$ is computed from Eq.~\eqref{eq:smd}, and the system is advanced using a semi-implicit Euler scheme:
\begin{align}
\dot{\mathbf{x}}_{\text{ref}}^{k+1}
&=
\dot{\mathbf{x}}_{\text{ref}}^{k}
+
\Delta t \, \ddot{\mathbf{x}}_{\text{ref}}^{k}, \\
\mathbf{x}_{\text{ref}}^{k+1}
&=
\mathbf{x}_{\text{ref}}^{k}
+
\Delta t \, \dot{\mathbf{x}}_{\text{ref}}^{k+1}.
\end{align}
Here, $k$ denotes the discrete control step, $\Delta t$ is the control period.
The resulting task-space reference is mapped to joint-space via an inverse kinematics (IK) solver~\citep{Zakka_Mink_Python_inverse_2025}, yielding joint position targets tracked by joint position controllers.

\subsection{VLM-based Planning}
\label{sec:vlm}
The first planning method uses vision foundation models and vision--language models to predict 3D contact points and normals, which are interpolated into a dense trajectory.

\textbf{Reasoning in Pixel Space.} Our pipeline first applies open-vocabulary segmentation using SAM3~\citep{carion2025sam} on rectified RGB images to localize the target object. For wiping tasks, candidate pixels are grid-sampled within the mask bounding box, while for drawing tasks, candidates are generated from predefined end-effector workspace rectangles. A vision--language model (VLM) is then prompted with the rectified image and the list of candidate pixel coordinates to output an ordered sequence of waypoints for each end-effector.

\textbf{Lifting to 3D Space.} Depth is estimated using Foundation Stereo~\citep{wen2025foundationstereo} and lifted to 3D using camera intrinsics. The resulting 3D point cloud is transformed from the camera frame to the world frame using the head pose and camera extrinsics. We then interpolate smooth trajectories at $50~\mathrm{Hz}$ with approach and retreat segments, pauses, separate free-space and contact speed limits. Contact forces $\mathbf{f}_\text{cmd}$ are applied along the estimated surface normals, while stiffness matrices $\mathbf{K}_p$ are computed from surface normals and fixed tangential and normal values, and damping matrices $\mathbf{K}_d$ are derived assuming critical damping. The pipeline is illustrated in Fig.~\ref{fig:planning}.

\subsection{Imitation Learning}
\label{sec:imitation}
\textbf{Data Collection.} We employ a teleoperation setup similar to~\citep{aldaco2024aloha}. A leader arm computes end-effector poses via forward kinematics in MuJoCo~\citep{todorov2012mujoco} and streams them to the follower, where the received pose is used to set the desired end-effector target $\mathbf{x}_{\text{des}}$. The follower then tracks this target using a low-level compliance controller during physical interaction.

\textbf{Policy Deployment.} We use a diffusion policy~\citep{chi2023diffusion} with a conditional 1D UNet that predicts diffusion noise over action sequences, conditioned on visual features extracted by a ResNet-18 encoder~\citep{he2016deep}. At each control step, the policy takes the observed end-effector pose $\mathbf{x}$ as input and outputs the desired end-effector target $\mathbf{x}_{\text{des}}$, which is passed to our compliance controller to generate  $\mathbf{x}_{\text{ref}}$ for execution.

\subsection{Model-based Planning}
\label{sec:model_based}
\textbf{Hybrid Force-Velocity Planning.} When contact location is known, a robust compliance profile can be computed in closed-form using optimally-conditioned hybrid servoing (OCHS)~\citep{hou2021efficient}. At each step, OCHS takes in the desired system velocities and contact information to compute a hybrid force-velocity control. The force control action maintains the safe engagement of desired contacts, while the velocity control action guarantees the desired object/robot velocities.
%
%
To implement the computed hybrid force-velocity action on our compliance controller, we assign high stiffness in the velocity-controlled directions, while setting low stiffness and force offset in the force-controlled directions for each end-effector.
The stiffness matrix for an end-effector with a desired Cartesian velocity $\bm{v}$ is computed as
\begin{equation}
\mathbf{K}
= k_{\mathrm{low}} \mathbf{I}
+ \left( k_{\mathrm{high}} - k_{\mathrm{low}} \right)
\frac{\bm{v}\bm{v}^{\top}}{\lVert \bm{v} \rVert^{2}} .
\end{equation}

To simplify implementation, the velocity component produced by OCHS is integrated over time to generate Cartesian position commands, which, together with the force and stiffness commands, are jointly executed by our compliance controller.
Using this formulation, the robot can perform contact-rich manipulation tasks, such as in-hand rotation, while maintaining robust contact behavior.

\section{Experiments}

\subsection{Hardware Setup}
We validate our compliance control method on four robotic platforms with distinct morphologies: ARX X5 robot arm with QDD motors; Unitree G1 with QDD motors; ToddlerBot~\citep{shi2025toddlerbot} with Dynamixel servos; and LEAP Hand~\citep{shaw2023leap} with Dynamixel servos. None of the platforms is equipped with force/torque sensors, and some even lack current sensing and control, making them ideal for evaluating our approach based solely on motor current or PWM feedback and position control.

\begin{figure}[t]
  \centering
  \includegraphics[width=\columnwidth]{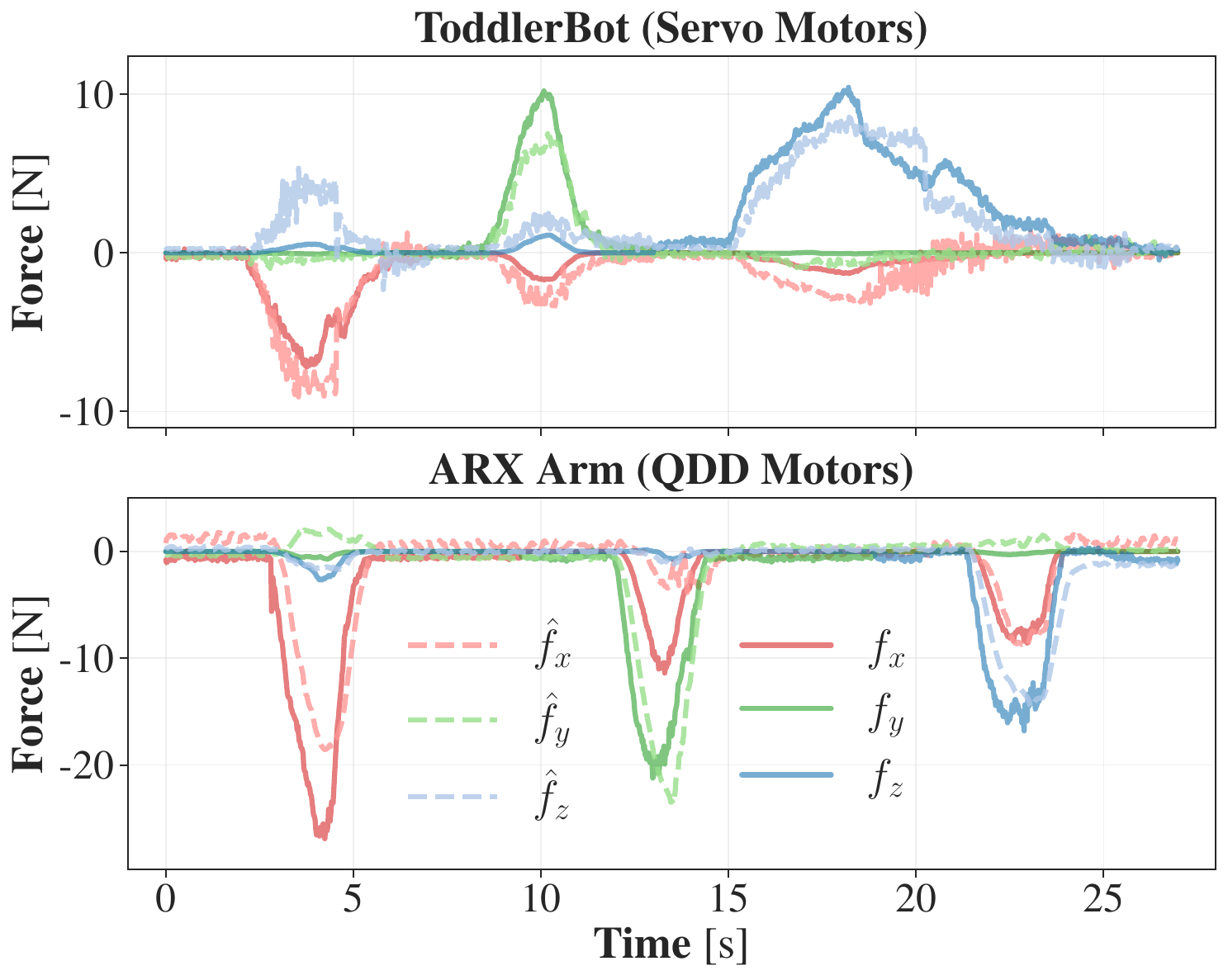}
  \caption{\textbf{Comparison with Force Sensor Readings.} Dashed lines ($\hat{f}$) denote the estimated forces, while solid lines ($f$) show ground-truth measurements from an ATI Mini45 sensor. The mean absolute error is $0.69 \pm 0.73~\mathrm{N}$ for ToddlerBot with servo motors (gear ratio $>200{:}1$) and $1.05 \pm 1.60~\mathrm{N}$ for ARX arm with QDD motor (gear ratio $\approx10{:}1$).}
  \vspace{-3mm}
  \label{fig:ati}
\end{figure}
\begin{figure}[t]
  \centering
  \includegraphics[width=\columnwidth]{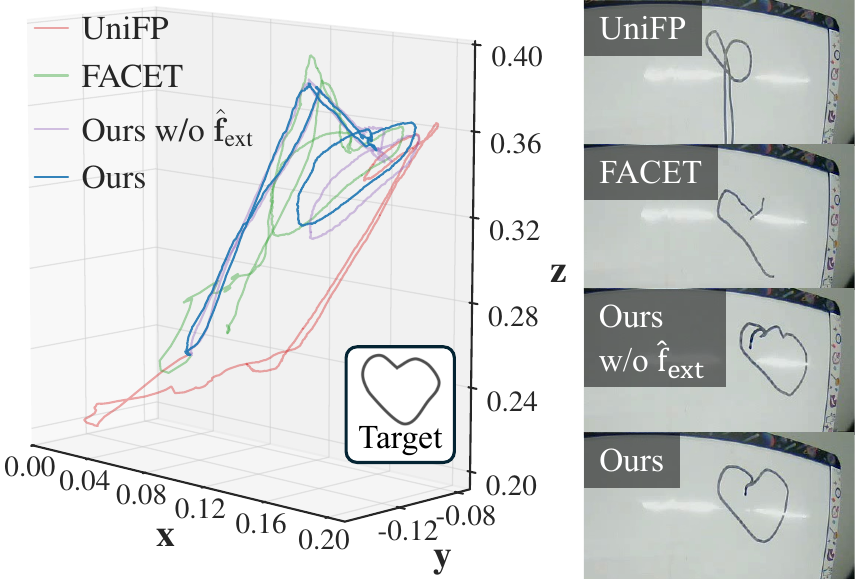}
  \caption{\textbf{Qualitative Comparison with Baselines.} In this experiment, ToddlerBot draws a heart on a whiteboard using a target trajectory generated by a VLM. All methods follow the same command, which specifies the desired end-effector position $\mathbf{x}_{\text{des}}$, velocity $\dot{\mathbf{x}}_{\text{des}}$, stiffness $\mathbf{K}_p$, damping $\mathbf{K}_d$, and commanded wrench $\mathbf{f}_{\text{cmd}}$. We visualize the 3D end-effector trajectories and real-world execution results for UniFP~\citep{zhi2025learning}, FACET~\citep{xu2025faceta}, our method without $\hat{\mathbf{f}}_{\text{ext}}$, and our full method.}
  \vspace{-2mm}
  \label{fig:baseline}
\end{figure}
\begin{table}[t]
\centering
\caption{\textbf{Quantitative Comparison with Baselines.} We report position tracking errors in millimeters and orientation tracking errors in radians, defined as
the deviation of $\mathbf{x}$ from
$\mathbf{x}_{\text{des}}$. We also report the humanoid’s root pitch as a proxy for contact force. Since the contact point is at an approximately fixed distance from the center of mass, the moment arm is nearly constant, making root pitch correlated with the applied force.}
\setlength{\tabcolsep}{5pt}
\begin{tabular}{@{}lccc@{}}
\toprule
    & Pos. Error~($\mathrm{mm}$) & Ori. Error~($\mathrm{rad}$) & Root Pitch~($\mathrm{rad}$) \\ \midrule
    UniFP~\citep{zhi2025learning} & 57.8 $\pm$ 30.2 & 0.147 $\pm$ 0.119 & 0.068 $\pm$ 0.020 \\
    FACET~\citep{xu2025faceta} & 22.4 $\pm$ 11.0 & 0.151 $\pm$ 0.087 & \textbf{0.018 $\pm$ 0.008} \\
    Ours w/o $\hat{\mathbf{f}}_\text{ext}$ & 22.5 $\pm$ 9.6 & 0.082 $\pm$ 0.040 & 0.060 $\pm$ 0.038 \\
    Ours & \textbf{15.9} $\pm$ \textbf{5.1} & \textbf{0.048} $\pm$ \textbf{0.043} & 0.029 $\pm$ 0.012 \\ \bottomrule
\end{tabular}
\vspace{-3mm}
\label{tab:baseline}
\end{table}
\begin{figure*}[p]
  \centering
  \includegraphics[width=\textwidth]{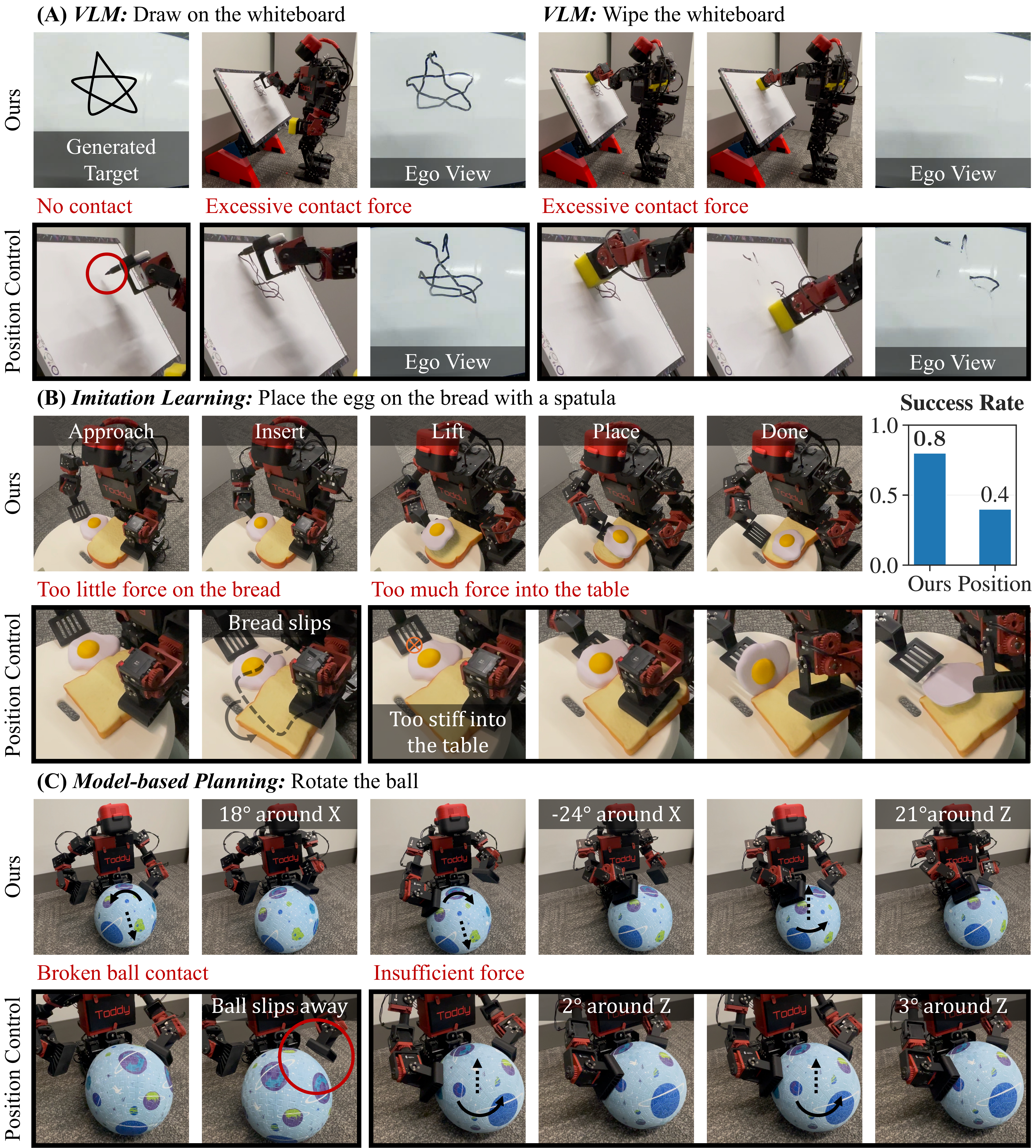}
  \caption{\textbf{ToddlerBot Results}. We demonstrate that Minimalist Compliance Control works on a floating-base humanoid robot and integrates seamlessly as a plug-and-play module across diverse high-level planners. Tasks include drawing and wiping on a whiteboard with a VLM-based policy, placing an egg on bread with a spatula using an imitation policy, and rotating a ball with a model-based policy. Odd rows (Ours) present successful executions of our method, while even rows (Position Control) illustrate representative failure cases of the baseline. Our controller maintains appropriate contact forces and stable interaction, whereas the position-control baseline frequently loses contact, applies insufficient force, or exerts excessive force, leading to task failure. Insufficient force typically fails to establish or maintain contact, while excessive force increases tangential friction and leads to larger tangential tracking errors.}
  \vspace{-3mm}
  \label{fig:toddy}
\end{figure*}

\subsection{External Wrench Estimation}
In this experiment, we press an ATI Mini45 force/torque sensor against the end-effector while running our compliance controller, apply pushes along each Cartesian axis, and estimate the resulting three-axis contact forces. The estimated forces closely match the sensor measurements, remaining near zero in free space and accurately tracking contact events. Quantitative results are reported in Fig.~\ref{fig:ati}.
Although friction is not explicitly modeled, accounting for the drive state $d$ (forward/backward) is sufficient for accurate estimation, with gearbox efficiency $\eta$ implicitly capturing friction and transmission losses from high gear reductions. To reduce noise, we apply an exponential moving average (EMA) filter with $\alpha=0.9$ for ToddlerBot and $\alpha=0.1$ for the ARX arm. A larger $\alpha$ yields more responsive but noisier estimates, highlighting the trade-off between responsiveness and smoothing introduced by the EMA filter.
A transient spike in ToddlerBot’s \(z\)-axis force around \(t=5\,\mathrm{s}\) arises from a temporarily ill-conditioned Jacobian that couples \(x\)- and \(z\)-axis estimates; in practice, this has negligible impact on task performance.

\subsection{Compliance Control}
\textbf{Baseline Implementation.} To evaluate our compliance controller against learning-based methods, we compare against two RL-based baselines, UniFP~\citep{zhi2025learning} and FACET~\citep{xu2025faceta}. In both cases, we retain the original RL policy formulations and implementations, while making a small number of implementation-level adjustments to ensure a fair comparison across methods.
The UniFP baseline directly uses the authors’ released codebase and policy architecture, while the FACET baseline is re-implemented to match the same infrastructure. 

During training, end-effector pose, external and commanded interaction wrenches, stiffness, and damping are uniformly sampled within the robot’s reachable workspace. The ranges ensure kinematic feasibility and remain consistent with those trajectories encountered during evaluation (Fig.~\ref{fig:baseline}), avoiding infeasible or out-of-distribution commands. We also adjust the reward weights to emphasize stable and responsive behavior under contact. More details are described in Appendix~\ref{sec:hyper}.

\textbf{Evaluation Results.} As shown in Fig.~\ref{fig:baseline}, the heart-shaped target trajectory is generated by the VLM-based policy and executed by all methods. UniFP and FACET struggle to track the position accurately while maintaining appropriate contact force with the whiteboard. The external wrench estimation in our method is also crucial for regulating contact force. Tab.~\ref{tab:baseline} reports quantitative results on position and orientation tracking errors, as well as the humanoid’s root pitch, which correlates with contact force. FACET exhibits the smallest root pitch because it fails to apply sufficient contact force to the board.

\textbf{Additional Attributes.} Our controller runs at about $12\,\mathrm{ms}$ per step, with about $10\,\mathrm{ms}$ spent in the IK solver~\citep{Zakka_Mink_Python_inverse_2025}. These timings are profiled on a Jetson Orin NX. Moreover, since motor currents respond to disturbances anywhere on the robot, the controller can react beyond end-effector contacts. Qualitative results are shown in the supplementary video.

\subsection{ToddlerBot Results}
 
We evaluate three high-level planners on the ToddlerBot humanoid platform to demonstrate that our compliance controller integrates seamlessly across planning paradigms. As shown in Fig.~\ref{fig:toddy}, we consider three tasks: drawing and wiping on a whiteboard with a VLM-based policy, placing an egg on bread with a spatula using an imitation policy, and rotating a ball using a model-based policy. These tasks require appropriate contact force, motivating the use of compliance control. Each task includes a position-control baseline, which fails due to loss of contact, insufficient force (failing to establish or maintain contact), or excessive force (increasing tangential friction and tracking errors). We observe that the quasi-static assumption does not degrade performance in these tasks.

\textbf{VLM-based Policy.} 
For drawing and wiping, we attach a sponge to ToddlerBot’s left hand and a pen to the right. Given a text prompt (e.g., ``star'') and an initial stereo observation of the whiteboard, vision foundation models and VLMs predict a 3D contact trajectory, which our compliance controller executes while maintaining contact force. After drawing, the robot captures a new observation to infer contact points for wiping. The compliance controller compensates for sometimes inaccurate predictions of 3D contact points and normals through compliant interaction. In contrast, the position-control baseline either fails to establish contact or applies excessive force, resulting in poor tangential tracking. The end-to-end pipeline runs in 3--6\,s, with about 90\% of the latency from the VLM API call. Qualitative results are shown in Fig.~\ref{fig:toddy}. More details are described in Appendix~\ref{sec:vlm_details}.

\textbf{Imitation Policy.}
For the egg placement task, we attach a spatula to ToddlerBot’s right hand. To train diffusion policies, we collect 200 demonstrations with the compliance controller active, logging both $\mathbf{x}_{\text{des}}$ and $\mathbf{x}_{\text{ref}}$. For a fair comparison, the compliance-enabled policy uses $\mathbf{x}_{\text{des}}$ as the action output, whereas the position-control baseline uses $\mathbf{x}_{\text{ref}}$, which is the trajectory after compliance-based adjustment via the spring--mass--damper model. Because inference-time trajectories are not strictly in-distribution, the compliance controller corrects local errors by reacting to estimated external wrenches, whereas a pure position controller cannot. The baseline often fails to stabilize the bread as a stopper or applies excessive force with the spatula, leading to poor tangential tracking. Our method achieves a success rate of $16/20=80\%$, compared to $8/20=40\%$ for the position-control baseline.

\textbf{Model-based Policy.} 
In the ball rotation task, we command a target angular velocity and follow a fixed sequence of rotation commands specifying the axis and which hand to use. For horizontal-axis rotations, the robot uses a single hand (left or right), whereas vertical-axis rotations are performed bimanually. The planner also takes the object's geometry and scale as input. Our method achieves an average rotation of $21^\circ$, while the baseline typically slips or rotates the ball by only $2^\circ$--$3^\circ$. Rotation is estimated from multi-view observations by fitting the ball contour, matching surface features, lifting them to 3D sphere points, and solving for the best-fit rotation.
    
\subsection{LEAP Hand Results}
\begin{figure*}[t]
  \centering
  \includegraphics[width=\textwidth]{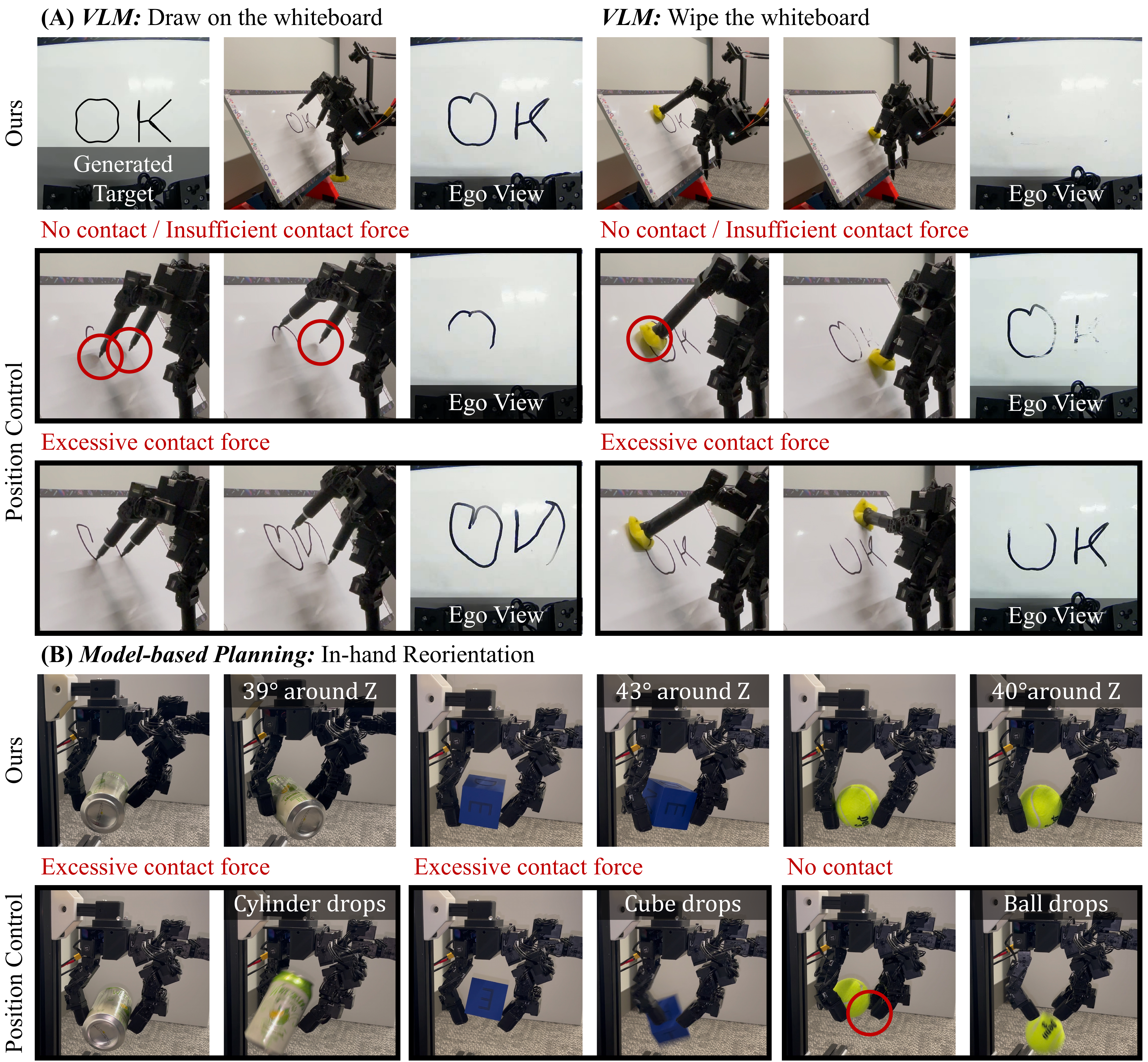}
  \caption{\textbf{LEAP Hand Results}. We show that Minimalist Compliance Control generalizes across embodiments, including dexterous manipulation. We evaluate two high-level strategies: a VLM-based policy for multi-finger drawing and wiping on a whiteboard, and model-based planning for in-hand reorientation of various objects. Both tasks demand precise contact force regulation. Each also includes a position-control baseline, which fails due to contact loss, insufficient force, or excessive force.}
  \vspace{-3mm}
  \label{fig:leap}
\end{figure*}

We further evaluate our controller on the LEAP Hand to show its effectiveness in dexterous manipulation. We consider two tasks: whiteboard drawing and wiping using a VLM-based policy, and in-hand reorientation of various objects using a model-based policy (Fig.~\ref{fig:leap}). Both require precise contact force regulation. Although the LEAP Hand’s mechanical design provides some passive compliance that a position controller can partially exploit, we show that incorporating active compliance control is still crucial for reliable performance.

\textbf{VLM-based Policy}. We use a setup similar to the ToddlerBot experiments, but the LEAP Hand can draw with multiple fingers simultaneously. A third-person view stereo camera observes the whiteboard. A pen is mounted on the index and ring fingers, and a sponge on the middle finger. Given the prompt ``Letter O and Letter K,'' a VLM-based policy predicts the 3D trajectory, which the compliance controller executes reliably. In contrast, the position-control baseline either fails to establish contact or tracks poorly due to excessive force.

\textbf{Model-based Policy}. For in-hand reorientation, we command target linear or angular velocities and execute a predefined sequence of commands. The planner receives the object’s geometry and scale as input. Our method achieves an average translation of $3.6\,\mathrm{cm}$ and a rotation of $27^\circ$, and can sustain reorientation for more than three cycles of translation and rotation commands. In contrast, the position-control baseline typically drops the object within the first few commands.
\section{Conclusion}

In conclusion, Minimalist Compliance Control achieves reliable compliance using widely available actuator signals, without force sensors or learning, making this capability accessible to many robotic systems with modern servo and QDD motors and lowering the barriers to contact-rich manipulation. We validate the method on a robot arm, a dexterous hand, and two humanoid robots across diverse contact-rich tasks, demonstrating that it generalizes across embodiments and integrates with vision--language, imitation, and model-based policies to achieve more stable contact than position-control baselines.

\textbf{Limitations.} 
Our current approach does not account for non-backdrivable or self-locking actuators. Motors with significant stiction or backlash are also not explicitly modeled and would require additional system identification. Additional effects, including joint acceleration, Coriolis terms, and thermal dynamics, could improve accuracy if modeled, but this introduces a trade-off between accuracy and complexity. In this work, we prioritize a minimalist design that provides sufficient accuracy for reliable performance in contact-rich tasks.

\IEEEpeerreviewmaketitle


\section*{ACKNOWLEDGMENT}
The authors would like to express their great gratitude to the Toddy Team for hardware support and the members of Stanford TML and REALab for helpful discussions.
This work was supported in part by the NSF Award \#2143601, \#2037101, \#2132519, \#2153854, Sloan Fellowship, and Stanford Institute for Human-Centered AI. 
The views and conclusions contained herein are those of the authors and should not be interpreted as necessarily representing the official policies, either expressed or implied, of the sponsors. 


\bibliographystyle{plainnat}
\bibliography{compliance}

\clearpage

\appendix
\subsection{RL Baseline Hyperparameters}
\label{sec:hyper}
\textbf{Input Commands Range.}
During training, we sample input commands within the ranges shown in Tab.~\ref{tab:baseline_command}; FACET and UniFP share the same command structure and ranges.
\begin{table}[h]
  \centering
  \caption{Command ranges to train UniFP and FACET.}
  \label{tab:baseline_command}
  \setlength{\tabcolsep}{5pt}
  \begin{tabular}{@{}l l@{}}
    \toprule
    \textbf{Command} & \textbf{Range} \\
    \midrule

    Left hand
      &
      $
      \begin{aligned}
        \mathbf{x}_{\text{des}} &\leftarrow \mathbf{0} \\
        \mathbf{f}_{\text{ext}} &\leftarrow \mathbf{0} \\
        \mathbf{f}_{\text{cmd}} &\leftarrow \mathbf{0}
      \end{aligned}
      $
      \\

    \midrule
    Right hand position $\mathbf{x}_{\text{des}}^{\text{pos}}$
          &
          $
          \begin{aligned}
            \mathbf{x} &\in [0.00, 0.10]~\mathrm{m} \\
            \mathbf{y} &\in [0.00, 0.09]~\mathrm{m} \\
            \mathbf{z} &\in [0.00, 0.15]~\mathrm{m}
          \end{aligned}
          $
          \\
    
    \midrule
    Right hand orientation $\mathbf{x}_{\text{des}}^{\text{rot}}$
          &
          $
          \begin{aligned}
            \mathbf{r_x} &\in [0.00, 0.10]~\mathrm{rad} \\
            \mathbf{r_y} &\in [0.2, 0.34]~\mathrm{rad} \\
            \mathbf{r_z} &\in [0.00, 0.00]~\mathrm{rad}
          \end{aligned}
          $
          \\

    \midrule
    Right hand $\mathbf{f}_{\text{ext}}$
      &
      $
      \begin{aligned}
        \mathbf{f}_{\text{ext},x} &\in [0, 5]~\mathrm{N} \\
        \mathbf{f}_{\text{ext},y} &\in [0, 3]~\mathrm{N} \\
        \mathbf{f}_{\text{ext},z} &\in [0, 3]~\mathrm{N}
      \end{aligned}
      $
      \\


    \midrule
    Right hand $\mathbf{f}_{\text{cmd}}$
      &
      $
      \begin{aligned}
        \mathbf{f}_{\text{cmd},x} &\in [0, 5]~\mathrm{N} \\
        \mathbf{f}_{\text{cmd},y} &\in [-1.5, 1.5]~\mathrm{N} \\
        \mathbf{f}_{\text{cmd},z} &\in [-2, 0]~\mathrm{N}
      \end{aligned}
      $
      \\

    \midrule
    Right hand $\mathbf{K}_{p}$ and $\mathbf{K}_{d}$
      &
      $
      \begin{aligned}
          \mathbf{K}_{p,x}^{\mathrm{pos}} &\in [120, 400] \\
          \mathbf{K}_{p,y}^{\mathrm{pos}} &\in [350, 400] \\
          \mathbf{K}_{p,z}^{\mathrm{pos}} &\in [350, 400] \\
          \mathbf{K}_{p}^{\mathrm{rot}} &\in [20, 40]^3 \\
          \mathbf{K}_d &= 2\,\mathbf{K}_p^{1/2}
        \end{aligned}
      $
      \\

    \bottomrule
  \end{tabular}
\end{table}

\textbf{Rewards.}
Table~\ref{tab:reward_terms} summarizes the reward terms to train
UniFP and FACET, together with their corresponding weights. Most weights differ from those in the original paper and are carefully tuned for our robot and tasks.
\begin{table}[h]
  \centering
  \caption{Reward terms and weights for UniFP and FACET.}
  \setlength{\tabcolsep}{5pt}
  \renewcommand{\arraystretch}{1.5}
  \begin{tabular}{l l r r}
    \toprule
    \textbf{Term} & \textbf{Definition} & \textbf{UniFP} & \textbf{FACET} \\
    \midrule
    eef\_site\_pos
      & $\exp\!\left(
          -\sigma_{\text{pos}}
          \lVert \mathbf{x}^{\text{pos}} - \mathbf{x}^{\text{pos}}_{\text{ref}} \rVert^2
        \right)$
      & 10.0 & 6.0 \\
    
    eef\_site\_vel
      & $\exp\!\left(
          -\sigma_{\text{linvel}}
          \lVert \dot{\mathbf{x}}^{\text{pos}} - \dot{\mathbf{x}}^{\text{pos}}_{\text{ref}} \rVert^2
        \right)$
      & 1.0 & 0.6 \\
    
    eef\_site\_rot
      & $\exp\!\left(
          -\sigma_{\text{rot}}
          \lVert \mathbf{x}^{\text{rot}} \ominus \mathbf{x}^{\text{rot}}_{\text{ref}} \rVert^2
        \right)$
      & 3.0 & 4.0 \\

    motor\_pos & $\exp(-\lVert q - q_{\text{ref}}\rVert^2)$ & 4.0 & 0.5 \\

    energy
      & $-\mathrm{mean}(\lVert\tau \dot{q}\rVert^2)$
      & 0.1 & 1.0 \\

    action\_rate
      & $-\mathrm{mean}(\lVert a_\text{t} - a_\text{t-1}\rVert^2)$
      & 0.08 & 2.0 \\

    collision & $\displaystyle -\sum \mathbf{1}(\lVert \mathbf{f}_{\text{contact}}\rVert > 0.1)$ & 3.0 & 3.0 \\
    \bottomrule
  \end{tabular}
  \label{tab:reward_terms}
\end{table}

\subsection{VLM Implmentation Details}
\label{sec:vlm_details}

\textbf{Prompt.} We use a structured prompt that specifies the task, target object, and candidate contact points (pixel coordinates) grouped by end-effector site, and require the model to return an ordered JSON sequence of waypoints per site (keyframes only; the controller later densifies the trajectory).

\begin{figure}[ht]
    \footnotesize
    \centering
    \begin{tcolorbox}[width=\linewidth, boxsep=0pt, left=4pt, right=4pt, top=4pt, bottom=4pt, sharp corners]

\textcolor{gray}{\textbf{Image:} \{image\}}

\textcolor{gray}{\textbf{Task:} \{task\_description\}}

\textcolor{gray}{\textbf{Target Object:} \{object\_label\}}

\textcolor{gray}{\textbf{Candidate Contact Points:}}\\
\textcolor{gray}{\{candidate\_lines\}}

\vspace{0.3em}
\textcolor{gray}{\textbf{Action Requirements:}}\\
\textcolor{gray}{Output an ordered list of waypoints (pixel coordinates) for each site to traverse the target region. The waypoint list should define a single continuous path that sweeps through the marked region without leaving any major sub-region uncovered. Focus on path topology—coverage order, direction changes, and sub-region transitions. We will densify the path later; your job is only to select the meaningful keyframes.}

\vspace{0.3em}
\textcolor{gray}{\textbf{Output JSON Format:}}\\
\textcolor{gray}{\{json\_lines\}}

\vspace{0.3em}
\textcolor{gray}{\textbf{JSON Rules:}}\\
\textcolor{gray}{Return valid JSON with double quotes.}\\
\textcolor{gray}{Maintain the order of points exactly for dynamic sequences.}\\
\textcolor{gray}{Do not include explanations.}

\vspace{0.3em}
\textcolor{gray}{\textbf{Begin Output:}}

    \end{tcolorbox}
    \vspace{-0.5em}
    \caption{\textbf{VLM prompt template.} For the wiping task, the task description is: ``Wipe up the black ink on the whiteboard using an eraser,'' where the default target object is ``black ink." For the drawing task, the task description is: ``Draw the \{object\_label\} on the whiteboard using a pen,'' where the target object corresponds to the provided \{object\_label\}.}
    \label{fig:vlm_prompt_wipe}
    \vspace{-3mm}
\end{figure}

\textbf{Candidate Generation.} For wiping, we segment the target object ``black ink" with SAM3 on the rectified left image, remove outliers using distance-to-center filtering, and generate a uniform grid of candidate pixels within the mask bounding box. For drawing, we estimate depth, project workspace rectangles using camera intrinsics and extrinsics, and sample a grid of candidate pixels within those regions.

\textbf{Floating-base Robots.} For floating-base robots, the controller continuously estimates torso orientation from the IMU. A PD stabilizer regulates torso pitch and yaw, issuing corrective commands at the hip and waist to maintain upright posture while the arms interact with the environment. During tasks such as drawing or wiping, the desired yaw is modulated by lateral hand displacement: as the hand moves sideways, the base yaws proportionally, with side-dependent limits, expanding the hand’s workspace while preserving balance. In addition to static standing balance, the supplementary video demonstrates that the VLM-based policy can also be integrated with a locomotion policy to wipe a vase.

\end{document}